# Deep Learning-Based Forecasting of Boarding Patient Counts to Address ED Overcrowding

Orhun Vural[1], Bunyamin Ozaydin[2,3], James Booth[4], Brittany F. Lindsey[5], Abdulaziz Ahmed[2,3,*]

## Abstract

**Objectives:** This study aims to develop deep learning models capable of predicting the number of patients in the emergency department (ED) boarding phase six hours in advance. The goal is to support proactive operational decision-making using only non-clinical, operational, and external contextual features, without relying on sensitive patient-level data.

**Materials and Methods:** Data were collected from five distinct data sources: ED tracking systems, inpatient census records, weather reports, federal holiday calendars, and local event schedules. Following comprehensive feature engineering, the data were aggregated at an hourly resolution, preprocessed to handle missing values and anomalies, and then merged into a unified dataset suitable for model training. Multiple time series deep learning models, including ResNetPlus, TSTPlus, and TSiTPlus (from the tsai library), were trained. Optuna, an automated hyperparameter optimization framework, was used to identify optimal hyperparameters for each model.

**Results:** The mean ED boarding count across the entire dataset was 28.7, with a standard deviation of 11.2. The best performance was achieved by the TSTPlus model, with a mean absolute error (MAE) of 4.30, mean squared error (MSE) of 29.47, root mean squared error (RMSE) of 5.43, and an $R^2$ score of 0.79. Model performance was also evaluated during extreme boarding periods, defined as hours when boarding counts exceeded one, two, or three standard deviations above the mean.

**Discussion:** This study demonstrates that accurate six-hour-ahead predictions of emergency department boarding counts are feasible using only routinely collected operational and contextual data. Evaluating multiple deep learning models across several dataset configurations revealed that expanding the input information generally improved forecasting accuracy, with the best results achieved by the TSTPlus model. Comprehensive feature engineering and systematic model evaluation enabled the identification of key input factors that most strongly influence boarding dynamics.

**Conclusion:** This study presents a deep learning-based framework for forecasting ED boarding counts six hours ahead without the need for patient-level clinical data. The approach offers a practical and generalizable solution for hospital systems, supporting proactive operational planning and contributing to efforts to mitigate ED overcrowding.

[1]Department of Electrical and Computer Engineering, University of Alabama at Birmingham, Birmingham, AL, USA
[2]Department of Health Services Administration, School of Health Professions, University of Alabama at Birmingham, Birmingham, AL, USA
[3]Department of Biomedical Informatics and Data Science, Heersink School of Medicine, University of Alabama at Birmingham, Birmingham, AL, USA
[4]Department of Emergency Medicine, University of Alabama at Birmingham, Birmingham, AL, USA
[5]Department of Patient Throughput, University of Alabama at Birmingham, Birmingham, AL, USA

* Corresponding Author, Abdulaziz Ahmed, PhD, Department of Health Services Administration, School of Health Professions, University of Alabama at Birmingham, Birmingham, AL 35233, United States, Email: aahmed2@uab.edu

## Introduction

Emergency Department (ED) overcrowding continues to be a significant challenge in hospital operations, negatively impacting patient outcomes, extending wait times, increasing healthcare costs, and even increasing violence against healthcare staff [1, 2]. A commonly used strategy for improving patient movement throughout hospitals is the Full Capacity Protocol (FCP), which has received recognition from the American College of Emergency Physicians (ACEP) as an effective framework for addressing operational challenges in emergency departments [3]. FCP serves as a structured communication framework between the ED and inpatient units and includes a tiered set of interventions aligned with varying levels of crowding severity. These intervention levels are triggered by specific Patient Flow Measure metrics (PFMs), which reflect the operational pressure within the ED. In literature, PFMs are sometimes referred to as Key Performance Indicators (KPIs) [4]. Key PFMs influencing ED overcrowding cover the entire patient journey—from initial registration to the end of the boarding period—and encompass factors both within the ED and in the broader hospital environment. Mehrolhassani et al. [4] conducted a comprehensive scoping review of 125 studies and identified 109 unique PFMs used to evaluate ED operations. The identified measures included key flow-related indicators such as treatment time, waiting time, boarding time, boarding count, triage time, registration time, diagnostic turnaround times (e.g., x-rays and lab results), etc. These represent just a subset of the many metrics used to monitor and improve ED performance. In this study, we focus specifically on the boarding process, which occurs when patients are admitted to the hospital but remain in the ED while awaiting an inpatient bed. This phase is widely recognized as one of the major causes of ED congestion, as boarded patients occupy treatment spaces and consume critical resources that could otherwise be used for incoming cases [5-7].

Boarding count refers to the number of patients who have received an admission decision—typically marked by an Admit Bed Request in the ED—but remain physically present in the ED while waiting for transfer to an inpatient unit [8]. Studies have investigated the impact of boarding on both patient outcomes and hospital operations. Su et al. [9] employed an instrumental variable approach to quantify the causal effects of boarding, reporting that each additional hour was associated with a 0.8% increase in hospital length of stay, a 16.7% increase in the odds of requiring escalated care, and a 1.3% increase in total hospital charges. Salehi et al. [10] conducted a retrospective analysis at a high-volume Canadian hospital and found that patients admitted to the medicine service had a mean ED length of stay of 25.6 hours and a mean time to bed of 15.9 hours. Older age, comorbidities, and isolation or telemetry requirements were significantly associated with longer boarding times, which in turn led to approximately a 0.9-day increase in inpatient length of stay after adjustment for confounders. Joseph et al. [11] found that longer ED boarding durations were independently associated with an increased risk of developing delirium or severe agitation during hospitalization, especially among older adults and those with dementia. Yiadom et al. [12] note that keeping admitted patients in the ED due to a lack of available inpatient beds puts considerable pressure on ED operations and is a leading cause of crowding and bottlenecks in patient movement. Boulain et al. [13] also reported that, in their matched cohort analysis, patients who experienced ED boarding times greater than 4 hours had a significantly higher risk of hospital mortality. Loke et al. [14] found that extended ED boarding is linked to high rates of verbal abuse toward staff—reported by 87% of nurses and 36% of attending physicians—and contributes to clinician burnout and dissatisfaction.

Several studies have focused on modeling or predicting ED boarding using both statistical and machine learning approaches. For example, Cheng et al. [15] developed a linear regression model to estimate the staffed bed deficit—defined as the difference between the number of ED boarders and available staffed inpatient beds—using real-time data on boarding count, bed availability, pending consults, and discharge orders. Hoot et al. [16] introduced a discrete event simulation model to forecast several ED crowding metrics, including boarding count, boarding time, waiting count, and occupancy level, at 2-, 4-, 6-, and 8-hour intervals. The model, which used six patient-level features, achieved a Pearson correlation of 0.84 when forecasting boarding count six hours into the future. However, this approach does not employ machine learning; it is based on simulation techniques rather than data-driven predictive modeling. More recent studies have leveraged machine learning to enable earlier predictions. Suley [17] developed predictive models to forecast boarding counts 1-6 hours ahead using multiple machine learning approaches, with Random Forest regression achieving the best performance. They demonstrated that when boarding levels exceed 60% of ED capacity, average length of stay increases from 195 to 224 minutes. However, these findings were based on agent-based simulation modeling rather than real hospital data, which may not reflect actual ED operations. Additionally, the study lacks detailed data descriptives such as mean and standard deviation of hourly boarding counts, limiting the ability to fully evaluate model performance. Kim et al. [18] used data collected within the first 20 minutes of patient arrival—including vital signs, demographics, triage level, and chief complaints—to predict hospitalization, employing five models: logistic regression, XGBoost, NGBoost, support vector machines (SVM), and decision trees. At 95% specificity, their approach reduced ED length of stay by an average of 12.3 minutes per patient, totaling over 340,000 minutes annually. Similarly, Lee et al. [19] modeled early disposition prediction using several algorithms, including multinomial logistic regression and multilayer perceptron, based on ED lab results and clinical features available approximately 2.5 hours prior to admission decisions.

Most existing models rely on a narrow set of input features, often limited to in-ternal ED data. In contrast, our study integrates a wider range of features, combining operational indicators with contextual variables such as weather conditions and significant events (e.g., holidays and football games), which originate outside of the hospital system. Importantly, our approach does not utilize patient-level clinical data but instead relies solely on aggregate operational data—structured, time-stamped numerical indicators that reflect system-level dynamics. This design simplifies data integration and enhances model generalizability across settings. Additionally, to our knowledge, there is no existing model that performs hourly boarding count prediction using real-world ED data by using deep learning models. Our results can be used to proactively inform FCP activation and serve as inputs to a clinical decision support system, enabling more timely and informed operational responses.

We develop a predictive model to estimate boarding count six hours in advance using real-world data from a partner hospital located in Alabama, United States of America, relying solely on ED operational flow and external features without incorporating patient-level clinical data. During this study, we worked closely with an advisory board consisting of representatives from various emergency department teams at our collaborating hospital. Based on recommendations from our advisory board and prior research identifying a six-hour window as critical for boarding interventions [17], we selected a six-hour-ahead prediction horizon. This timeframe provides sufficient lead time for operational decision-making and initiating FCP steps. This approach enables a shift from reactive to proactive management in the ED. This proactive approach supports real-time decision-making to help mitigate ED overcrowding before critical thresholds are reached. As part of our contribution, we perform extensive feature engineering to derive key flow-based variables—including treatment count, waiting count, waiting time, and treatment time—that are not directly observable in the raw data. We also construct and evaluate multiple datasets with different combinations of these features to identify the most effective input configuration for prediction. Finally, we implement deep learning models with automated hyperparameter optimization using Optuna [20], which enables dynamic definition of search spaces and efficient trial pruning. To enhance model interpretability, we analyze attention matrices from transformer models to visualize and understand which temporal patterns and input features most influence predictions. Moreover, by avoiding sensitive patient-level inputs, our method simplifies data collection and improves generalizability, making it readily adaptable for use across diverse hospital systems.

## Materials and Methods

The methodological workflow of this study comprises several key stages designed to develop a reliable model for predicting boarding count. The process begins with the collection of multi-source data that captures both internal ED operations and external contextual factors. Next, feature engineering is applied separately to each data source, with a focus on creating patient flow metrics to derive informative variables. Following preprocessing steps such as cleaning, categorization, and normalization, all data sources are merged into a unified dataset at an hourly resolution. Using this unified dataset, five distinct datasets are created based on different feature combinations, as shown in Table 2. In the subsequent stage, model training is carried out using three deep learning architectures: Time Series Transformer Plus (TSTPlus) [21], Time Series Inception Transformer Plus (TSiTPlus) [22], and Residual Networks Plus (ResNetPlus) [23].

Following training, model evaluation is performed to assess predictive performance across the constructed datasets using standard regression metrics. The primary objective is to predict the boarding count six hours in advance, supporting proactive decision-making to mitigate ED overcrowding. Additionally, performance under high-volume boarding conditions was examined through extreme case analysis, as shown in Table 3, to assess the model's reliability during periods of elevated crowding. To further enhance interpretability, an attention analysis was conducted using the internal attention weights from the transformer architecture as shown in Figure 2. This explainability module reveals which temporal patterns and input features most strongly influence the model's predictions.

### Data Source

To accurately predict boarding counts, five distinct data sources were utilized to construct a comprehensive dataset capturing internal ED operations and external contextual factors relevant to ED dynamics. These sources include: (1) ED tracking, (2) inpatient records, (3) weather, (4) federal holiday, and (5) event schedules. The same data sources were also used in our previous study [24], which focused on predicting another PFM—waiting count. All data were processed and aligned to an hourly frequency, resulting in a unified dataset with one row per hour. The dataset spans from January 2019 to July 2023, providing over four years of continuous hourly records to support model development and evaluation.

The ED tracking data source captures patient movement throughout the ED, starting from arrival in the waiting room to transfer to an inpatient unit or discharge from the ED. Each patient visit is linked to a unique Visit ID, with 308,196 distinct visits included in the data source. The data provides timestamps for location arrivals and departures, enabling reconstruction of patient flow over time. It also includes room-level information that indicates whether the patient is in the waiting area or the treatment area during their stay. The Emergency Severity Index (ESI) is recorded using standardized levels from 1 to 5, along with a separate category for obstetrics-related (OB) cases. Clinical event labels indicate the type of care activity at each step (e.g., triage, admission, examination, inpatient bed request), while clinician interaction types identify the provider involved (e.g., nurse or physician).

The inpatient dataset contains hospital-wide records of patients admitted to inpatient units, independent of their ED status. Each record includes timestamps for both arrival and discharge, allowing accurate calculation of hourly inpatient census across the entire hospital. By aligning these timestamps with the study period, the number of admitted patients present in the hospital at any given hour can be determined. A total of 293,716 unique Visit IDs are included in this data source, forming the basis for constructing a continuous hospital census variable used in the prediction models.

The weather dataset was obtained from the OpenWeather History Bulk [25] and includes hourly observations collected from the meteorological station nearest to the hospital. This external data source provides environmental context not captured by internal hospital systems. It includes both categorical weather conditions—such as Clouds, Clear, Rain, Mist, Thunderstorm, Drizzle, Fog, Haze, Snow, and Smoke—and one continuous variable: temperature. These features were used to enrich the dataset with relevant temporal environmental information.

The holiday data source captures official federal holidays observed in the United States, obtained from the U.S. government's official calendar [26]. Each holiday was mapped into the dataset at an hourly resolution by marking all 24 hours of the holiday with a binary indicator variable. This representation allows the model to incorporate the presence of holidays consistently across the entire study period.

The event data source includes football game schedules for two major NCAA Division I teams located in nearby cities close to our partner hospital. Game dates were collected from the teams' official athletic websites [27, 28] and added to the dataset by marking all 24 hours of each game day with a binary indicator. This information was incorporated to provide additional temporal context associated with scheduled local events.

Table I presents a summary of all data sources and shares the descriptive analysis of the features derived from them. This table reflects the structure and content of the dataset following the completion of all preprocessing and feature engineering steps.

**Feature Engineering and Preprocessing**

Raw data underwent feature engineering and preprocessing to construct meaningful features from multiple sources. Several aggregated flow metrics were derived to reflect hourly inpatient and ED dynamics, enabling the development of predictive models based on structured, time-aligned data.

Feature engineering was primarily applied to the ED tracking and inpatient datasets, from which operational metrics were calculated to describe patient flow across different areas of the hospital.

From the ED tracking data, nine distinct aggregated flow metrics were engineered to capture hourly patient activity within the department: (1) boarding count, representing the number of patients in the boarding phase, calculated from the time an Admit Bed Request is placed until the patient checks out of the ED; (2) boarding count by ESI level, grouping boarding counts into three ESI categories (1&2, 3, and 4&5) to assess boarding distribution by acuity; (3) average boarding time, measuring the mean duration of boarding per hour in minutes; (4) waiting count, indicating how many patients were in the waiting room during each hour; (5) waiting count by ESI level, breaking down the total waiting count into the same three ESI categories to reflect differences in waiting room congestion by acuity; (6) average waiting time, reflecting the mean duration patients spent waiting; (7) treatment count, denoting the number of patients in treatment rooms each hour; (8) average treatment time, representing the typical treatment duration during those periods; and (9) an extreme case indicator was created as a binary variable to flag hours in which a selected flow metric exceeded its historical mean plus one standard deviation. From the inpatient data, a single feature—hospital census—was engineered to capture the number of admitted patients present in the hospital at each hour. The feature list used in this study includes more than nine features, as detailed in Table 1. However, feature engineering was specifically applied to these nine features because they are not directly available in the raw dataset and must be created through additional calculations.

To prepare the data for modeling, several preprocessing steps were applied, including the following:

- Categorical weather conditions were grouped into five broader categories—Clear, Clouds, Rain, Thunderstorm, and Others—based on their semantic similarity to simplify downstream modeling. Specifically, 'Clouds' and 'Mist' were grouped under Clouds; 'Rain' and 'Drizzle' under Rain; 'Thunderstorm' remained as Thunderstorm; and 'Fog,' 'Haze,' 'Snow,' and 'Smoke' were combined under Others. Weather data for 5 hours was missing on different days in March 2023; these gaps were imputed using the weather observations from the preceding hour.
- To improve data quality and ensure consistency, specific steps were taken to address missing values and remove unrealistic records. For missing values in the Emergency Severity Index (ESI) field—approximately 2% of the dataset—a value of 3 was assigned, as ESI level 3 accounted for nearly 60% of all recorded entries. In addition, 601 visits with ESI values not among the common categories of 1, 2, 3, 4, 5, or OB (such as records labeled as "staff") were removed. Visits with waiting times exceeding 9 hours were excluded, representing 2.1% of the data, since 90% of these cases were extreme outliers with durations spanning several months. Additionally, 51 visits were removed where patients remained in the treatment room for more than seven months with identical treatment-leaving timestamps, indicating likely system logging errors. Furthermore, if a visit had a boarding start time recorded as later than the corresponding boarding end time—an operationally implausible scenario likely caused by logging issues—the visit was excluded; this criterion resulted in the removal of 6 visits. Finally, 74 visits with recorded boarding times longer than 300 hours were excluded, 70 of which had identical checkout timestamps and were also likely caused by data entry or logging issues.
- Lagged and rolling features were computed using a custom function that systematically transformed each selected variable by generating lagged versions and rolling averages. For each variable, lag features were created by shifting the original values backward by 1 to N hours, producing a series of lagged inputs corresponding to different historical time steps. This enables the model to learn from recent historical values. To capture local trends and smooth out noise, rolling mean features were calculated using a centered moving average over a specified window size.

- Given the unusual operational conditions during the early stages of the COVID-19 pandemic, data from April 2020 to July 2020 were excluded, as boarding patterns during this period did not reflect routine processes. As shown in Appendix 1, the monthly average boarding counts and times were notably lower than in other years, likely due to the uncertainty and disruption experienced by individuals and healthcare institutions at that time.

After completing the feature engineering and preprocessing steps, all data sources were merged into a unified dataset with an hourly resolution. An hourly timeline was created from 2019-01-01 09:00:00 to 2023-07-01 20:00:00, totaling 37,236 time points. This timeline served as the basis for aligning and merging the five data sources, with each row representing one hour. Table I presents the descriptive analysis of the final dataset, summarizing the distribution of both engineered features and raw variables across the full study period. This includes statistical details for numerical variables (mean, standard deviation, and range), percentages for categorical features, and event counts for binary indicators. The resulting dataset provides a structured and comprehensive foundation for predictive modeling of boarding count and other ED-related dynamics.

Table 1: Data Description Table

| Feature | Date Range, Average ± Standard Deviation (Range) for Numerical Features, % for Categorical Features, and Event Counts |
|---|---|
| Date Range | |
| Year | 5 years |
| Month | 12 Months |
| Day of Month | Days 1-31 |
| Day of Week | 7 Days |
| Hour | 24 Hours |
| Boarding Count (Target Variable) | 28.7± 11.2 (0 – 73) |
| Boarding Count by ESI Levels | |
| ESI levels 1&2 | 17.1 ± 7.6 (0 – 52) |
| ESI level 3 | 11.4 ± 5.5 (0 – 37) |
| ESI levels 4&5 | 0.1 ± 0.4 (0 – 5) |
| Average Boarding Time | 621 ± 295.8 (0 – 2446) (minutes) |
| Waiting Count | 18 ± 10 (0 – 65) |
| Waiting Count by ESI Levels | |
| ESI levels 1&2 | 4.7 ± 3.9 (0 – 24) |
| ESI level 3 | 10.6 ± 6.7 (0 – 46) |
| ESI levels 4&5 | 2.5 ± 2.2 (0 – 18) |
| Average Waiting Time | 86.7 ± 62.9 (0 – 425) (minutes) |
| Treatment Count | 53.9 ± 11.7 (5 – 98) |
| Average Treatment Time | 502.8 ± 196 (71 – 1643) (minutes) |
| Extreme Case Indicator | 6361 rows |
| Hospital Census | 788 ± 75 (421 – 1017) |
| Temperature | 62.84 ± 15.55 (8.3 – 100) °F |
| Weather Status | |
| Clouds | 60.1% |
| Clear<br>Rain<br>Thunderstorm<br>Others | 22.9%<br>15.45%<br>1.22%<br>0.4% |
| Football Game 1 | 54 Games |
| Football Game 2 | 49 Games |
| Federal Holidays | 46 Days |

## Dataset Construction

Following the completion of feature engineering and preprocessing, five distinct datasets were constructed, each representing a different combination of features. Each dataset has the shape (samples, features, 1), where lags and all other operational and contextual variables act as individual features. These variations were intentionally designed to assess the impact of specific feature groups—such as flow metrics, weather conditions, hospital census, and temporal indicators—on model performance. The main goal was to identify which feature combinations provide the most accurate boarding count predictions and best detect extreme cases, as detailed in the Extreme Case Analysis section. Each dataset was separately used to train and test the models, enabling direct comparison of results across configurations.

Table 2: Summary Statistics of Hourly Features Used in Model Training and Testing

| **Data Sources and Scaling** | **Features** | **Lags and Rolling Mean*** | **DS1** | **DS2** | **DS3** | **DS4** | **DS5** |
|---|---|---|---|---|---|---|---|
| ED Tracking | Boarding Count (Target) | Lags (W=12) | X | X | X | X | |
| | | Lags (W=24) | | | | | X |
| | | Rolling Mean (W=4) | | | | X | X |
| | Average Boarding Time | No Lags | | X | X | X | |
| | | Lags (W=12) | | | | | X |
| | Treatment Count | No Lags | | X | X | X | |
| | | Lags (W=12) | | | | | X |
| | Waiting Count | No Lags | | X | X | X | |
| | | Lags (W=12) | | | | | X |
| | Boarding Count by ESI Levels | | | | | X | X |
| | Waiting Count by ESI Levels | | | | | X | X |
| | Average Treatment Time | | | X | X | X | X |
| | Average Waiting Time | | | X | X | X | X |
| | Extreme Case Indicator | | | X | X | X | X |
| | Year, Month, Day of the Month, Day of the Week, Hour | | X | X | X | X | X |
| Inpatient | Hospital Census | No Lags | | X | X | X | |
| | | Lags (W=12) | | | | | X |
| Weather | Weather Status (5 Categories) | | | X | X | X | X |
| | Temperature | | | | X | X | X |
| Holiday | Federal Holiday | | | | X | X | X |
| Events | Football Game 1 | | | | X | X | X |
| | Football Game 2 | | | | X | X | X |

## Model Architectures and Training

Three time series deep learning models—TSTPlus, TSiTPlus, and ResNetPlus—were used to forecast ED boarding count. These models were implemented using the tsai library [29], a PyTorch [30] and fastai-based framework [31] tailored for time series tasks such as forecasting, classification, and regression. The models were selected based on their strong performance in our previous study [24], which focused on predicting ED waiting count.

TSTPlus is inspired by the Time Series Transformer (TST) architecture [32] and utilizes multi-head self-attention mechanisms to capture temporal dependencies across input sequences. The model is composed of stacked encoder layers, each containing a self-attention module followed by a position-wise feedforward network. These components are equipped with residual connections and normalization steps to enhance training stability and performance. Model explainability is based on attention scores, highlighting which time steps and features most influence predictions.

TSiTPlus is a time series model inspired by the Vision Transformer (ViT) architecture [33], designed to improve the modeling of long-range dependencies in sequential data. It transforms multivariate time series into a sequence of patch-like tokens, enabling the model to process the input in a way similar to how ViT handles image patches. The architecture incorporates a stack of transformer encoder blocks, each composed of multi-head self-attention layers and position-wise feed-forward networks, with optional features such as locality self-attention, residual connections, and stochastic depth.

ResNetPlus is a convolutional neural network (CNN) designed for time series forecasting, utilizing residual blocks to capture hierarchical temporal features across multiple scales. Each block applies three convolutional layers with progressively smaller kernel sizes (e.g., 7, 5, 3), combined with batch normalization and residual connections to promote stable training and effective deep feature learning. This architecture enables efficient extraction of both short- and long-term patterns from sequential data.

The dataset was partitioned into training (70%), validation (15%), and testing (15%) subsets for all three deep learning models. Hyperparameter optimization was conducted using Optuna for the TSTPlus, TSiTPlus, and ResNetPlus models. Optuna enables dynamic construction of the hyperparameter search space through a define-by-run programming approach. The framework primarily uses the Tree-structured Parzen Estimator (TPE) [34] for sampling but also supports other algorithms such as random search and CMA-ES [35]. To improve search efficiency, Optuna implements asynchronous pruning strategies that terminate unpromising trials based on intermediate evaluation results. The number of optimization trials is defined by the user to control the overall search budget. For each of these models, 50 trials were conducted to explore hyperparameters including learning rate (ranging from 1e-4 to 0.1), dropout rate (0.0 to 0.3), weight decay (0.0 to 0.2), optimizer (Adam [36], SGD [37], Lamb [38]), activation function (relu, gelu, silu), batch size (32 or 64), number of fusion layers (128 to 512, with a step size of 64), and training epochs.

**Evaluation**

Model performance was evaluated on the test set using the following metrics:

- Mean Absolute Error (MAE): Measures the average size of the errors.
- Mean Squared Error (MSE): Gives more weight to larger errors.
- Root Mean Squared Error (RMSE): The square root of MSE, in the same unit as the target.
- $R^2$ Score: Shows how well the predictions match the actual values.

The model was also evaluated in extreme cases. An hour was classified as "Extreme" if the boarding count was 40 or higher (mean plus one standard deviation), "Very Extreme" if it was 51 or higher (mean plus two standard deviations), and "Highly Extreme" if it was 62 or higher (mean plus three standard deviations). The distribution of the boarding count and the classification thresholds are provided in Appendix 2.

**RESULTS**

Following model training and evaluation, performance metrics were computed for each model–dataset combination using four standard measures: MAE, MSE, RMSE, and $R^2$ score. As illustrated in Figure 1, these metrics summarize model performance across all datasets.

The best result was achieved by the TSTPlus model trained on Dataset 3, yielding an MAE of 4.30, MSE of 29.66, RMSE of 5.44, and $R^2$ of 0.79 on the test set. This configuration was selected through automated hyperparameter search using Optuna, which systematically explores the parameter space and can identify sensitive, non-integer values—such as a learning rate of 0.0246, dropout rate of 0.1335, and weight decay of 0.0542—that yield optimal model performance. The most effective combination also included 200 epochs, the Lamb optimizer, and relu activation function. The TSTPlus model architecture consists of a stack of three transformer encoder layers, each incorporating multi-head self-attention, residual connections, normalization, and position-wise feed-forward networks. After feature extraction, the output is passed through a dense fusion layer with 128 neurons and relu activation to aggregate temporal features before the final prediction. Dropout is applied to the fusion layer to reduce overfitting, and weight decay further helps to regularize the model by penalizing large weights. The model is trained using MSE loss.

When evaluating the three different models across different datasets, variations in performance were observed depending on the feature set used, as shown in Figure 1. As shown in Table 2, Dataset 1 serves as our baseline, including only the target variable's lagged values along with standard temporal features such as year, month, day, and hour. As a result, all models performed worst on Dataset 1 due to limited features, with MAE/MSE of 4.77/35.13 (TSTPlus), 4.82/36.73 (ResNetPlus), and 5.57/48.84 (TSiTPlus). For Dataset 2, adding features such as average boarding time, treatment count, waiting count, average treatment time, average waiting time, extreme case indicator, hospital census, and weather status improved performance, reducing MAE to 4.51 (TSTPlus), 4.69 (ResNetPlus), and 5.31 (TSiTPlus). In Dataset 3, in addition to the features in Dataset 2, temperature, federal holiday, and two separate football game indicators were added as additional contextual features. This further improved model performance, with MAE decreasing to 4.30 for TSTPlus, 4.44 for ResNetPlus, and 5.24 for TSiTPlus. In Dataset 4, additional features were included for boarding count and waiting count, separated by three different ESI levels (ESI 1&2, ESI 3, and ESI 4&5). However, with these additions, model performance did not improve, and error rates increased for all models. In Dataset 4, additional features were included for boarding count and waiting count, separated by three different ESI levels (ESI 1&2, ESI 3, and ESI 4&5). However, these additions did not improve model performance, and error rates increased for all models, with MAE values of 4.68 for TSTPlus, 4.62 for ResNetPlus, and 5.28 for TSiTPlus. Finally, in Dataset 5, different lag windows were applied to various features. Although TSiTPlus showed a decrease in error rate, the best results for TSTPlus and ResNetPlus remained with Dataset 3.

Figure 1: Performance metrics (MAE, MSE, RMSE, R²) across datasets using deep learning models.

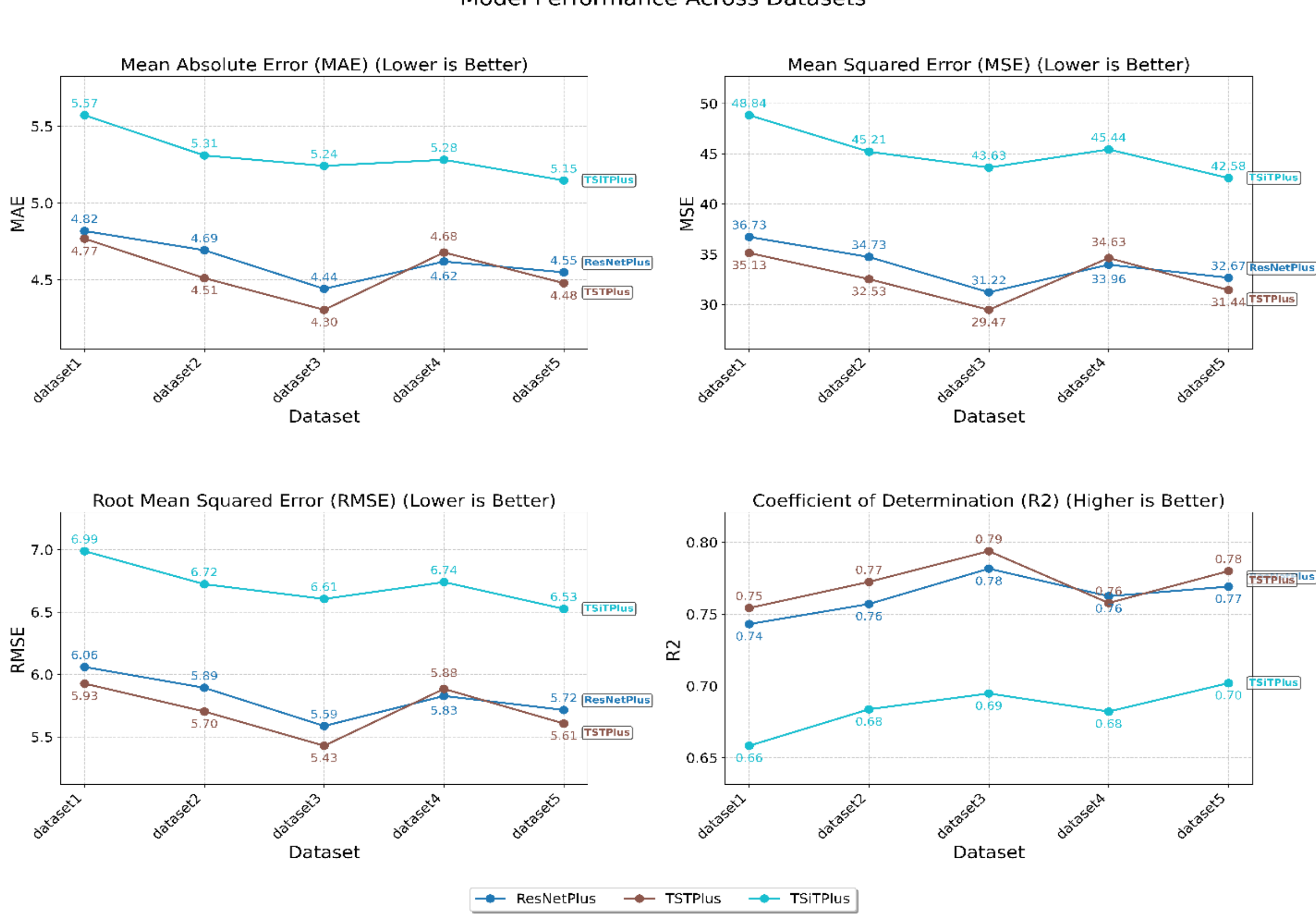

## Extreme Case Analysis

Extreme case analysis aims to identify periods when boarding counts significantly exceed typical levels, indicating unusual strain on hospital resources. To define these cases, boarding count data were categorized using hourly mean (28.7) and standard deviation (11.2) values, as shown in Table 1. As shown in Appendix B.1, boarding counts were classified into four groups: Normal (<40),

Extreme (≥40), Very Extreme (≥51), and Highly Extreme (≥62). Thresholds were determined using one, two, and three standard deviations above the mean boarding count, denoted as 1σ, 2σ, and 3σ, respectively, providing a structured approach to detect and quantify extreme overcrowding events.

Table 3 shows the prediction performance of TSTPlus across all datasets under extreme case scenarios. TSTPlus was selected for this analysis because it achieved the best overall results among the models. This evaluation is important for understanding how well the model performs during periods of severe ED crowding and for identifying which dataset provides the most reliable predictions in extreme situations. As the table indicates, Dataset 3 consistently produced the lowest MAE and RMSE values at all extreme thresholds, highlighting its strong predictive performance during high boarding counts. These results suggest that the features included in Dataset 3 are especially effective for forecasting ED overcrowding under challenging conditions. Notably, for Dataset 3, the MAE for the entire test dataset was 4.30, while the MAE for the "Extreme" scenario (boarding count ≥40) was slightly lower at 4.24. This demonstrates that the model maintains—and even slightly improves—its prediction accuracy during periods of high crowding, further confirming its robustness in critical situations.

Table 3. Performance of the TSTPlus model, reported as MAE and RMSE, under different extreme case scenarios for each dataset.

| Dataset | Mean + 1σ<br>Extreme (≥ 40) | Mean + 2σ<br>Very Extreme (≥ 51) | Mean + 3σ<br>Higly Extreme (≥ 62) |
|---|---|---|---|
| | MAE / RMSE | MAE / RMSE | MAE / RMSE |
| Dataset1 | 4.85 / 6.01 | 6.92 / 7.95 | 11.70 / 12.40 |
| Dataset2 | 4.34 / 5.45 | 5.79 / 6.89 | 10.03 / 10.95 |
| Dataset3 | **4.25 / 5.35** | **5.31 / 6.44** | **8.88 / 9.95** |
| Dataset4 | 4.61 / 5.73 | 6.60 / 7.59 | 11.46 / 12.05 |
| Dataset5 | 4.82 / 5.97 | 6.71 / 7.76 | 11.26 / 12.14 |

## Model Explainability

To better understand how individual features and their lagged values influence the boarding count predictions, model explainability analysis was conducted for TSTPlus, the best-performing model in this study. For this purpose, a new dataset was specifically constructed for explainability analysis. In this dataset, all features were represented with their respective lagged values across a 12-time-step window, resulting in an input shape of (samples, timesteps, features). In this data structure, each feature has a lag value equal to its sequence length. This design allows the attention mechanism within TSTPlus to capture both the temporal dependencies and the relative importance of each feature at different time points. The goal was to identify which operational, contextual, and temporal variables contribute most to the model's predictions, and to interpret the relative importance of lagged historical information.

This analysis utilized the internal self-attention mechanisms of the TSTPlus transformer model. During inference, attention weights were extracted from all encoder layers and averaged across heads, layers, and samples to quantify the importance of each input feature and its lagged values. Feature group importance was then computed by aggregating the total attention received by each feature and its lags, which enabled assessment of the variables with the greatest impact on the model's predictions, including boarding count, hospital census, treatment count, weather conditions, and special event indicators, as shown in Figure 2. The Figure 2 reveals that boarding count itself is the most critical predictor, which is clinically intuitive as current boarding levels strongly influence future patterns due to ED overcrowding persistence. Temporal features demonstrate significant predictive power, with month and day showing substantial importance, suggesting strong seasonal patterns in boarding. Patient flow metrics form the core predictive capability, as treatment time, treatment count, and boarding time all rank highly, indicating current operational efficiency directly impacts future scenarios. Weather conditions show moderate influence, with temperature being most predictive, while external events like football games have minimal predictive value for 6-hour ahead predictions.

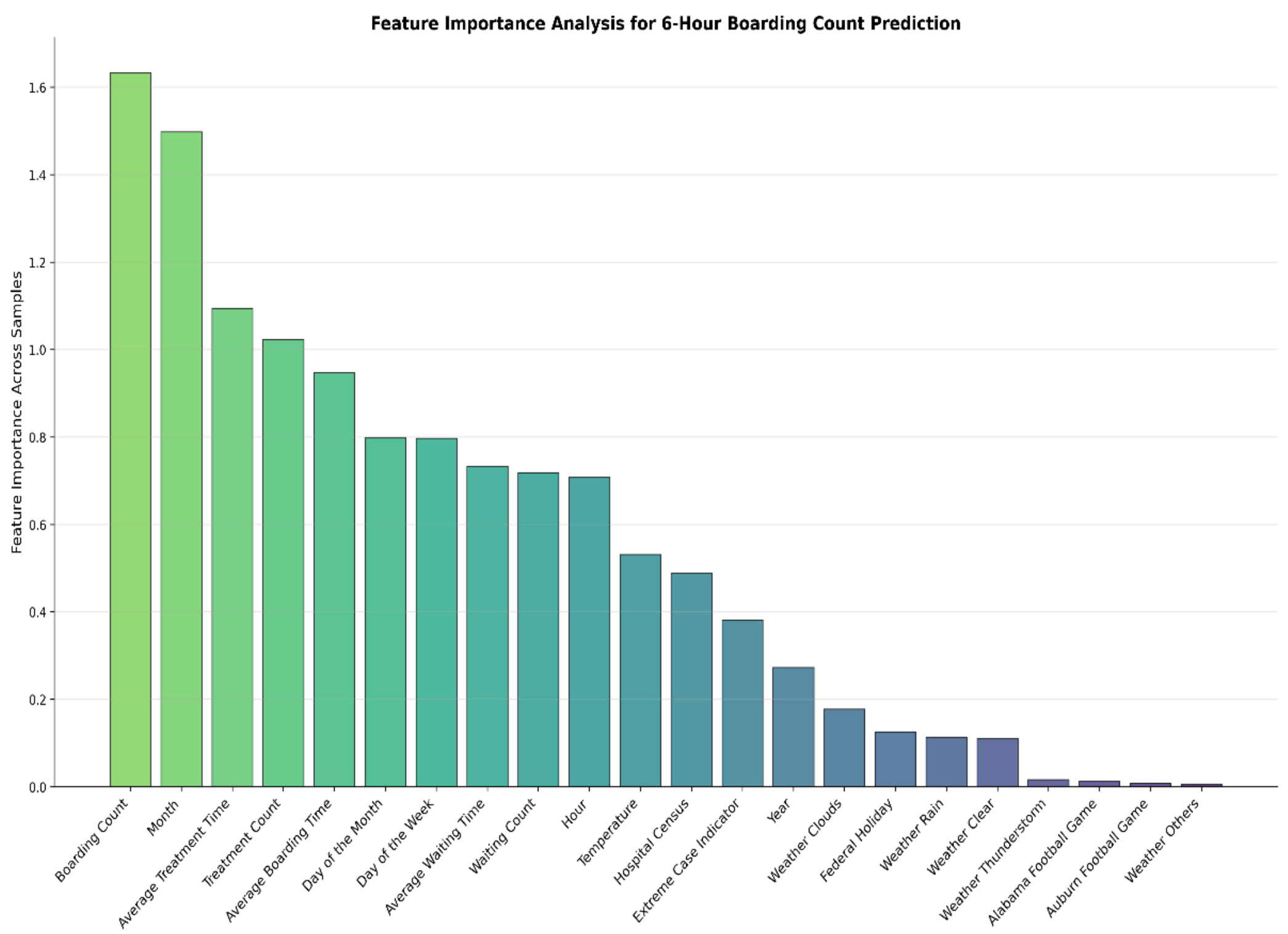

Figure 2. Feature importance for TSTPlus with input data shaped as (samples, timesteps, features)

Figure 3 displays the temporal lag attention patterns from the TSTPlus transform-er model for 6-hour boarding count predictions. The analysis reveals that the current time point (t-0) receives the highest attention, indicating present conditions are most critical for predictions. Recent historical data (t-1 and t-2) also demonstrate high at-tention weights, while mid-range lags (t-3 through t-6) show moderate attention and distant periods (t-7 through t-10) receive lower scores. Notably, the most distant time points (t-11 and t-12) show increased attention, potentially capturing daily cyclical patterns or shift-change effects that influence boarding dynamics.

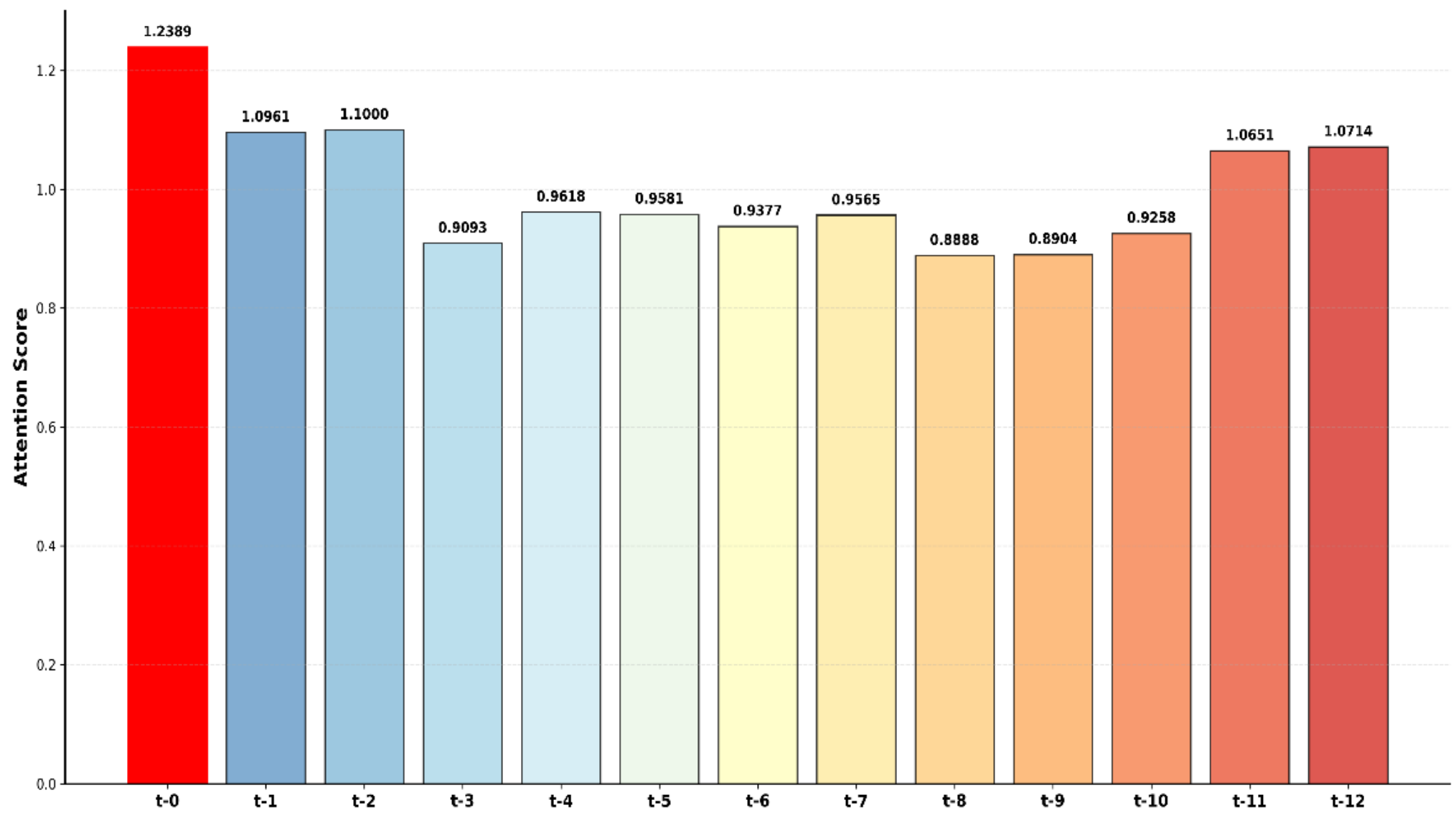

Figure 3. Temporal Lag Attention Patterns for 6-Hour Boarding Count Prediction

## DISCUSSION

This study demonstrates the practical value of deep learning–based short-term forecasting for ED boarding counts, supporting proactive operational management. Accurate six-hour-ahead predictions are especially useful for enabling timely activation of the FCP giving hospital administrators the ability to plan staff assignments, adjust bed allocations, and organize patient transport before crowding becomes critical. This proactive approach has the potential to mitigate ED congestion and reduce operational strain.

Several key contributions distinguish this work. Extensive feature engineering was performed on operational data, generating new flow metrics such as treatment count, waiting count, waiting time, and treatment time, and systematically constructing lagged and rolling features to better capture temporal and operational trends. Multiple dataset configurations were evaluated, demonstrating the advantage of combining operational and contextual variables for robust forecasting. Automated hyperparameter optimization using Optuna ensured that each deep learning model achieved optimal performance by systematically searching the parameter space. By accurately predicting an important Patient Flow Measure—boarding count—this approach directly supports proactive FCP activation and provides an operational foundation for better coordination with inpatient units, timely allocation of inpatient beds, and the potential to activate additional inpatient surge capacity in response to predicted crowding.

Among the models evaluated, TSTPlus achieved the best predictive performance, with an MAE of 4.30 and an $R^2$ of 0.79 on the test set. Attention-based explainability analysis revealed that current boarding count, temporal features, and patient flow metrics had the greatest influence on the model's predictions, while weather and special events contributed less to short-term forecasts—reflecting the dominant role of recent operational dynamics in ED crowding. The feature importance results (Figure 2) further confirmed that boarding count, month, day, and patient flow measures are the most critical predictors, whereas weather features (notably temperature) and external events were of moderate or minimal value for short-term prediction.

Extreme case analysis provided additional insights into model robustness under high-demand conditions. By categorizing boarding counts according to established statistical thresholds, the study assessed model accuracy during "Extreme," "Very Extreme," and "Highly Extreme" crowding scenarios. TSTPlus consistently maintained, and even slightly improved, its predictive accuracy during these critical periods, suggesting strong reliability for real-world application.

There are several limitations to note. Although TSTPlus includes learnable positional encodings, transformer models do not inherently capture directionality or causality as directly as RNNs or CNNs, which may be a disadvantage for tasks with strict temporal ordering. Furthermore, while this research demonstrates the feasibility of using predictive models to trigger FCP, the actual benefits of such proactive management were not evaluated in simulation or operational settings and should be assessed in future studies. The work relies on aggregate operational data, not patient-level records, which reduces privacy concerns and avoids the complexities of clinical data integration. This approach supports broad generalizability and potential adoption in other ED environments.

## CONCLUSION

Boarding count is a critical patient flow measure and a major driver of emergency department (ED) overcrowding, which can negatively impact patient care, increase wait times, and strain hospital resources. This study addresses the need for timely and accurate forecasting of boarding counts by developing deep learning models using real-world emergency department data from a partner hospital. In our dataset, the boarding count had a mean of 28.7 and a standard deviation of 11.2 per hour, providing context for evaluating model performance. Leveraging comprehensive feature engineering, multiple dataset configurations, and automated hyperparameter optimization, the TSTPlus model achieved the best results, with a mean absolute error of 4.30 and an $R^2$ of 0.79 on the test set. Accurate six-hour-ahead predictions of boarding count are important because they enable hospital administrators to anticipate and manage crowding before critical thresholds are reached, supporting proactive FCP activation, effective staff and bed management, and inpatient surge planning. The proposed framework does not rely on patient-level clinical data, which facilitates generalizability and privacy protection, making it readily applicable to diverse hospital environments. Future research should focus on evaluating the real-world impact of these forecasts through simulation or operational studies and integrating these models into clinical decision support systems to further improve ED crowding management.

## ACKNOWLEDGMENT

This project was supported by the Agency for Healthcare Research and Quality (AHRQ) under grant number 1R21HS029410-01A1.

### Conflicts of Interest

The authors declare no conflicts of interest.

## Data Availability

The data used in this study was provided by our partner hospital under strict confidentiality and IRB regulations and cannot be shared due to privacy agreements. Access is limited to authorized researchers for approved purposes only. The source code and model implementation are available at https://github.com/drorhunvural/ED_OverCrowding_Predictions.

## APPENDIX

### Appendix 1

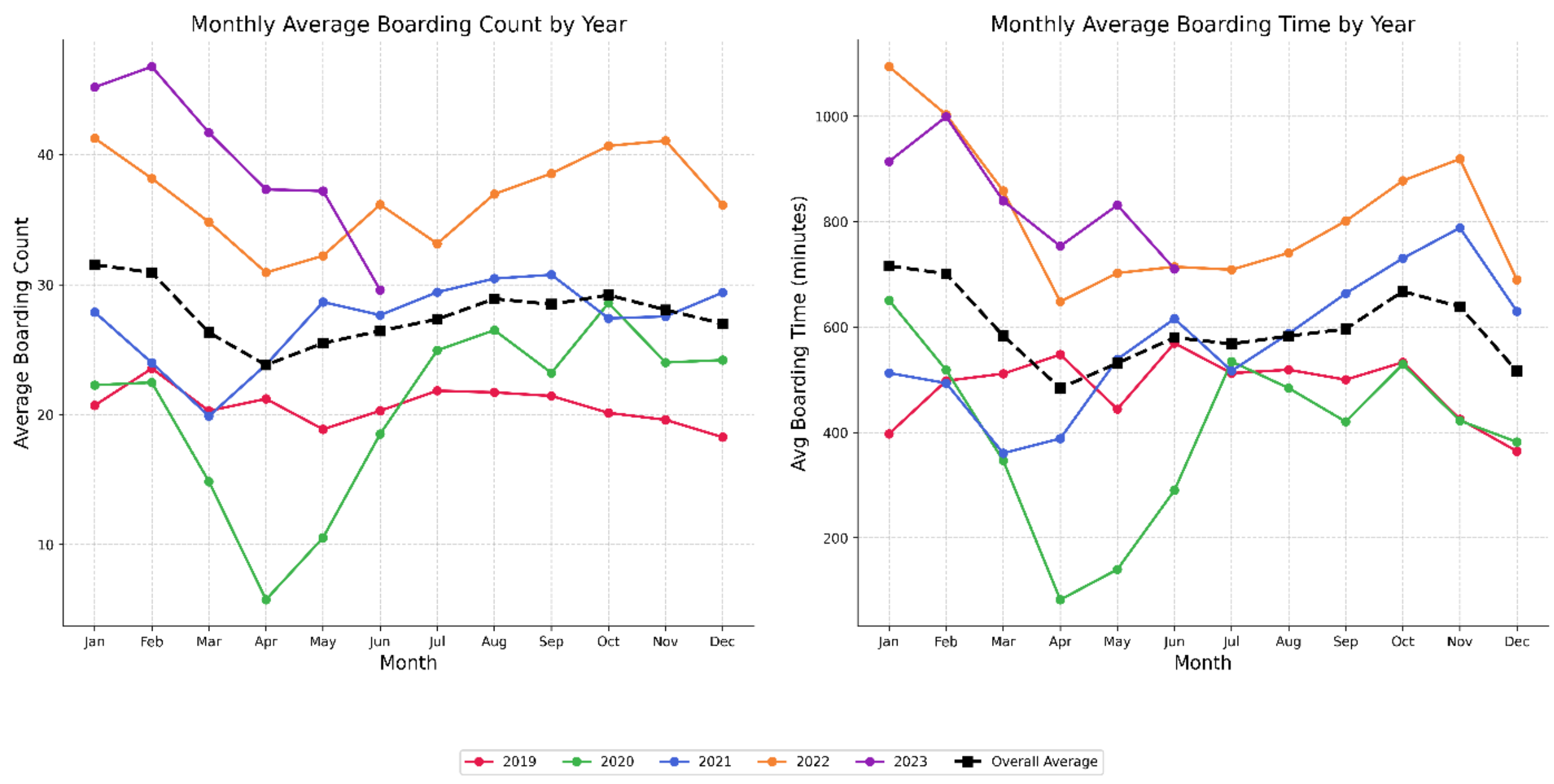

### Appendix 2

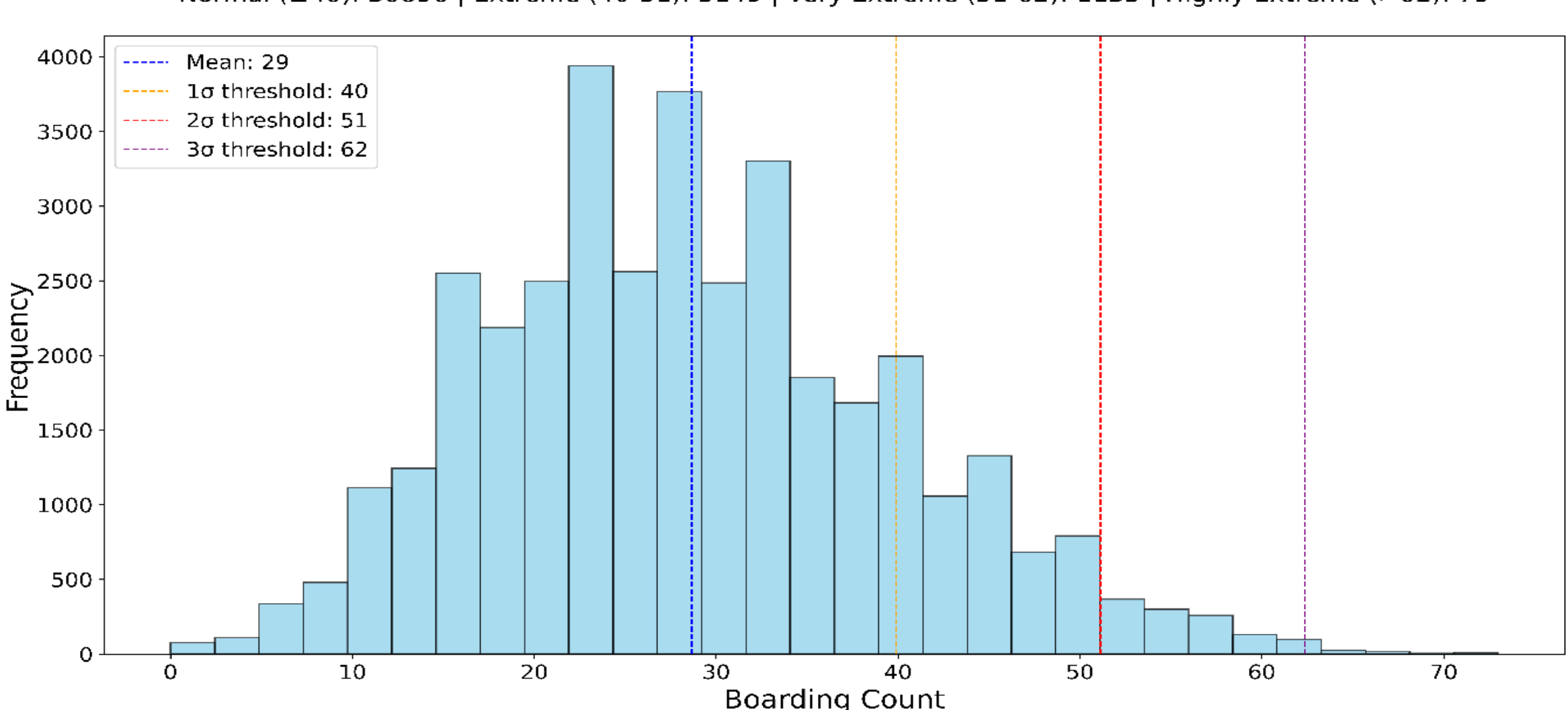